\documentclass[letterpaper, 10 pt, journal, table]{IEEEtran}  

\IEEEoverridecommandlockouts                              

\usepackage{graphics} 
\usepackage{float}
\usepackage{cite}
\usepackage{adjustbox}
\usepackage{mathptmx} 
\usepackage{amsmath} 
\usepackage{amssymb}  
\usepackage{todonotes}
\usepackage{multirow}
\usepackage{algorithm}
\usepackage{algpseudocode}
\makeatletter
\let\NAT@parse\undefined
\makeatother
\usepackage[colorlinks, linkcolor=green, citecolor=blue]{hyperref}

\def\BibTeX{{\rm B\kern-.05em{\sc i\kern-.025em b}\kern-.08em
  T\kern-.1667em\lower.7ex\hbox{E}\kern-.125emX}}
  
\title{\LARGE \bf
DaFoEs: Mixing \underline{Da}tasets towards the generalization of vision-state deep-learning \underline{Fo}rce \underline{Es}timation in Minimally Invasive Robotic Surgery
}

\author{Mikel De Iturrate Reyzabal$^{1}$, Mingcong Chen$^{2}$, Wei Huang$^{2}$, Sebastien Ourselin$^{1}$ and Hongbin Liu$^{1, 2}$ 
\thanks{*This work was supported by UK Research and Innovation in the Engineering and Physical Sciences Research Council 
        (EPSRC) iCASE [2446549] and the Centre for Artificial Intelligence and Robotics, Hong Kong Institute of Science \& Innovation, Chinese Academy of Sciences (InnoHK).}
\thanks{$^{1}$Mikel De Iturrate Reyzabal, Sebastien Ourselin and Hongbin Liu are with the School of Biomedical Engineering and Imaging Sciences,
        King's College London, UK, $^{2}$Wei Huang, Mingcong Chen and Hongbin Liu are with the Centre for Artificial Intelligence and Robotics Hong Kong Institute of Science \& Innovation, Chinese Academy of Sciences, Hong Kong.}%
\thanks{Correspondence: {\tt mikel.de\_iturrate\_reyzabal@kcl.ac.uk}}
\thanks{The code used in this research for the training and validation of the NN models can be found at: https://github.com/mikelitu/DaFoEs}
}

\begin{document}

\maketitle
\thispagestyle{empty}
\pagestyle{empty}

\begin{abstract}

Precisely determining the contact force during safe interaction in Minimally Invasive Robotic Surgery (MIRS) is still an open research challenge. Inspired by post-operative qualitative analysis from surgical videos, the use of cross-modality data driven deep neural network models has been one of the newest approaches to predict sensorless force trends. However, these methods required for large and variable datasets which are not currently available. In this paper, we present a new vision-haptic dataset (DaFoEs) with variable soft environments for the training of deep neural models. In order to reduce the bias from a single dataset, we present a pipeline to generalize different vision and state data inputs for mixed dataset training, using a previously validated dataset with different setup. Finally, we present a variable encoder-decoder architecture to predict the forces done by the laparoscopic tool using single input or sequence of inputs. For input sequence, we use a recurrent decoder, named with the prefix R, and a new temporal sampling to represent the acceleration of the tool. During our training, we demonstrate that single dataset training tends to overfit to the training data domain, but has difficulties on translating the results across new domains. However, dataset mixing presents a good translation with a mean relative estimated force error of $5\%$ and $12\%$ for the recurrent and non-recurrent models respectively. Our method, also marginally increase the effectiveness of transformers for force estimation up to a maximum of $\simeq 15\%$, as the volume of available data is increase by $150\%$. In conclusion, we demonstrate that mixing experimental set ups for vision-state force estimation in MIRS is a possible approach towards the general solution of the problem.

\end{abstract}

\section{Introduction}
\label{sec:intro}

Minimally Invasive Robotic Surgery (MIRS) has been a broad area of research in recent years. One of the main challenges of these procedures is the safe and precise interaction with the soft environment, in order to avoid tissue damage or excessive bleeding. One way of increasing the {safeness of the operations} is to {exchange }control and simulate the current forces during the surgical task with real-time haptic feedback. Moreover, haptic feedback has been demonstrated to effectively reduce the amount of force during mock training soft tissue phantoms under different experimental set ups\cite{bahar_surgeon-centered_2020, abiri_multi-modal_2019}. Generating trustworthy and accurate haptic feedback is still an open research challenge\cite{patel_haptic_2022}. One of the solutions for this problem is the use of force sensors at the tip of the robotic tool, to measure the contact force. There are various types of sensors depending on the mechanical or physical property that translates into the force.  However, the development of such technologies is a difficult task as the list of requirements is extended. Some of the most relevant characteristics of these sensors are scalability, exact wide working range and biocompatibility, to name but a few. 

Other approaches to estimate the interaction force with the environment required the use of mathematical or physical models. These models can be adapted to the different robotic systems, such as the Da Vinci\cite{pique_dynamic_2019} Robotic system (Intuitive Surgical), to estimate the interaction wrench at the end effector, using the dynamic modelling of the string base robotic forceps. However, these {methods are constrained to these unique systems}. {On the other hand, specific environments could be modelled using the }Finite Element Method (FEM) {approach}. {In which,} {Lagrangian} Constraint Equations {are used} to determine the contact forces {from the tool}. Although there are many viable simulators available for this task\cite{talbot_surgery_2015}, the {solution of such equations is} largely influenced by the {geometry and physical parameters, as well as, the modelling of} constraints.

{In the absence of sensors and models} surgeons {use visual information to qualitatively determine the effectiveness of a given motion and, therefore, infer the contact force. Although this kind of methods have been used to quantitatively score the performance of certain tasks\cite{gao_jhu-isi_nodate}, they present a clear subjective individual bias.} {In order to find a more general solution} based on these visual performance evaluations, Deep Neural Networks have {become the standard} approach for vision-based sensorless force estimation. {Specifically}, Convolutional Neural Networks (CNN) presented a light and fast approach to the solution of this problem, but as demonstrated in some research visual predictions are sensitive to unseen environments\cite{chua_toward_2021}. Consequently, in the same study Chua et al. show that the combination of visual and robot state increases the robustness and generalization of such models.

\begin{figure*}[t]
    \centering
    \includegraphics[width=0.95\textwidth, height=10.1cm]{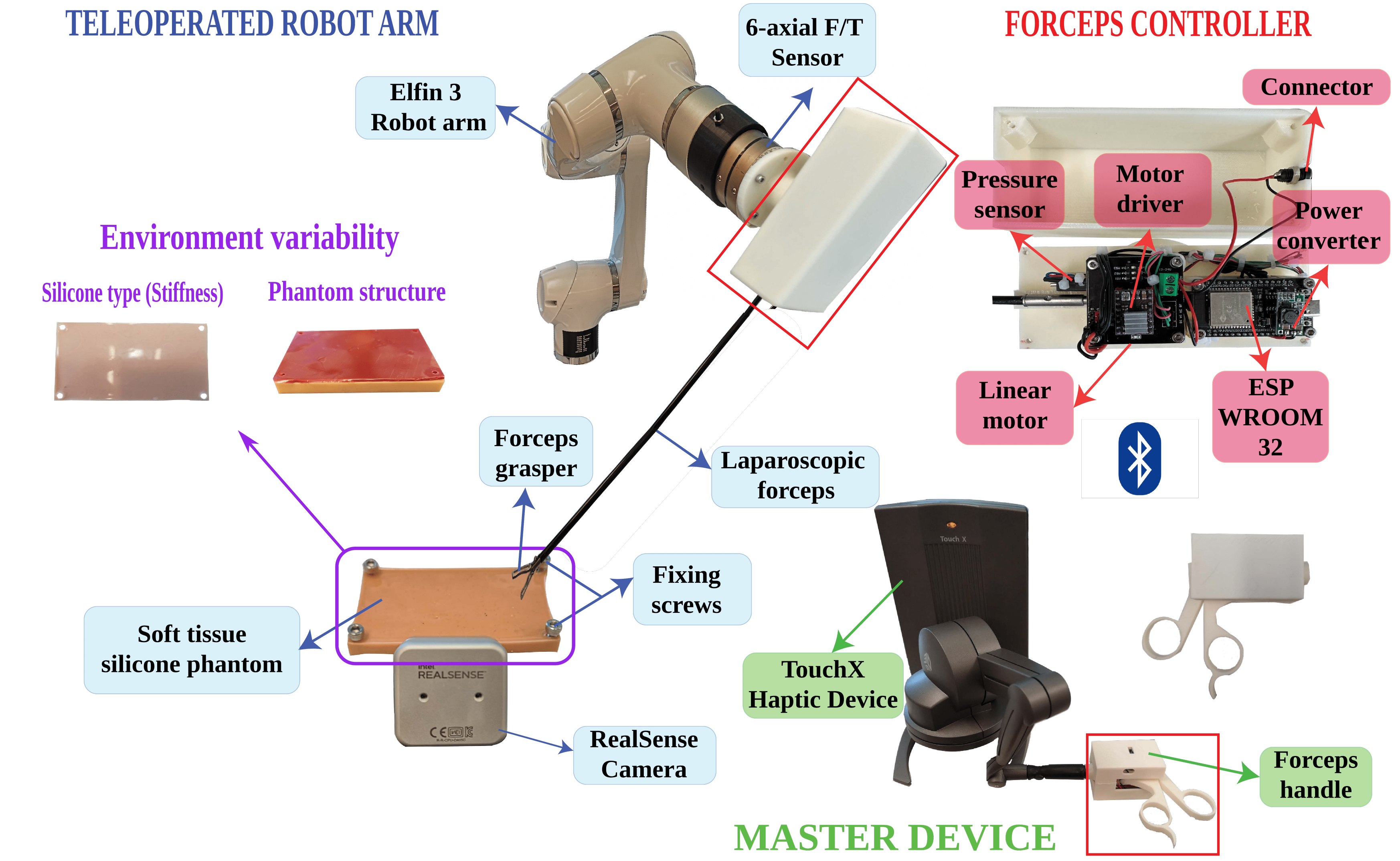}
    \caption{Complete experimental setup used for the collection of the DaFoEs  (\underline{Da}taset for \underline{Fo}rce \underline{Es}timation) dataset. The setup is divided into 3 main components, color coded: teleoperated robot arm (blue), master controller (green) and the forceps controller (red). In the left side of the image, we show the different possibilities for the soft tissue environment.}
    \label{fig:full_setup}
\end{figure*}

{However}, CNNs {, due to their lack of temporal expansion, miss relevant information regarding the force and} usually require from heavier robot state vectors. Here is where Recurrent Neural Networks (RNN) shown better results for lighter robot state vectors and multiple camera views\cite{marban_recurrent_2019, liu_camera_2022, sabique_stereovision_2022}. {More specifically in surgery, newest research have concluded that the combination of convolutional image encoder and recurrent decoder showed better results even for the most complex scenarios \cite{sabique_data_2024, sabique_investigating_2023}. Nevertheless,} in other vision-based research areas, such as classification, Vision Transformers (ViT) \cite{dosovitskiy_image_2021}, coming from Natural Language Processing problems, have shown more promising results than convolutional architectures. ViTs, due to the use of fully connected layers they do have longer back-propagation times for a correct parameter optimization than CNNs, but usually required reduced training iterations and are more sensible to data augmentation functions. However, there has been work around the possibility of accelerating the transformers, reducing the number of patches analyse in every level based on the attention score\cite{fayyaz_adaptive_2022}.

In this research, {we present a new dataset for vision based sensorless force estimation (DaFoEs) using} a teleoperated laparoscopic forceps {mounted} onto a robot arm controlled by a customized commercial haptic master device to include the actuation of the forceps grasp. {The main contributions of this manuscript are: (1) create a pipeline to generalize visual-state inputs of deep neural network training for sensorless force estimation from different data stream; (2) present a new neural network architecture by combining a ViT based image encoder and a recurrent decoder with a specific temporal window and (3) the comparison of such model with previous work on this area.\cite{chua_toward_2021}}

\section{{Experimental setup \& Data collection}}
\label{sec:mat}

In this section, {we discuss the different components that creates our teleoperated robot-tissue interaction setup.} {Additionally, we present} {the {process of} visual-state data acquisition}

\subsection{{Robot teleoperation}}
\label{sec:setup}

{In order to analyse the adaptability of the model to different structures and stiffness, we created different soft tissue silicone phantoms over a general rectangular geometry of size $122 \times 72$ mm and thickness $13$ mm. In addition, for the double layer structure {we crafted a} second thin phantom of the same size and $3$ mm thickness. Regarding the stiffness, we {used} two different types of silicone with variable tensile strengths Ecoflex 00-30 ($1.38$ MPa) and DragonSkin20 ($3.79$ MPa)}. {We screwed} the silicone phantom {to the working table on its edges and placed on the centre of the camera view.} The {Realsense D405 (Intel Company) stereo-}camera was placed 10 cm away from the phantom {and fixed using an adjustable tripod. For this paper, we only used the RGB sensor from the camera.} 

{The teleoperated unit is formed by an Elfin3 Collaborative Robot (Han's Robot Company). To mimic the conditions from laparoscopy, we mounted a commercial laparoscopic forceps as the end effector using a 3D printed PLC control box attachment. Additionally, for the force recordings we placed a M3815A1 6 axis F/T sensor (Sunrise Instruments) between the robot original end effector and the printed box.} {For a more realistic control of the forceps}, we designed a new handle {that substitutes the end cap from the} Touch X (3DSytems) haptic device. {The new handle contains a PCB board (ESP32-WROOM) which is connected via Bluetooth to the board on the control box. This actuates the motor to open and close the forceps{, linearly mapped with the values from a linear pressure sensor}. For a more detailed view of the control unit and the entire set up refer to Fig.} \ref{fig:full_setup}. 

{We use a {follower-leader} teleoperation framework to guide the forceps tip positionusing commands from the haptic device. In order to constrain the movement of the robot to a small workspace ($150 \times 100 \times 30$ mm), we limit {the range of the angle for each joint} using inverse kinematics equation. We update the robot configuration at $200$ Hz using commands from the haptic device {sent} via TCP/IP protocol.}

\subsection{DaFoEs: Dataset for Force Estimation acquisition}
\label{sub:dafoes}

We design different experiments of interaction with the soft phantom tissue {where we} applied force into a variety of points with variable approaching angles. In order to increase the variability even more, we {move} the phantom {across} 5 different positions separated 1cm from each other. This resulted in a total of 90 clips of around 30 seconds each, making a total of $\simeq70,000$ frames and $540,000$ robot state readings. 

For each of the different repetitions of the teleoperation experiments we {record the} visual and robot state {data}. {The former visual data was recorded at a frequency of 30 Hz. The latter state was recorded at a frequency of 200Hz which is the same as the controller step time mentioned in Sec. \ref{sec:setup}. The robot state vector has 26 elements for every time step, as represented in Eq.} \ref{eq:rob-state}.

\begin{equation} 
\label{eq:rob-state}
    \rho_{i} = [p_{E_{i}}, o_{E_{i}}, J_{robot_{i}}, p_{H_{i}}, o_{H_{i}}, J_{H_{i}}]
\end{equation}

where $p_{E_{i}}$ \& $o_{E_{i}}$ are the {position (Cartesian 3DoF) and orientation (quaternions 4DoF)} of the robot end effector, $J_{robot_{i}}$ is the 6 joint position for the robot arm; $p_{H_{i}}$ and $o_{H_{i}}$ are the position and orientation for the haptic device handle and $J_{H_{i}}$ the position of the joints of the haptic device.

{Additionally we record the 3 axial Cartesian force (Fx, Fy and Fz) from the force sensor. In order to align the force axis with the {end effector frame,} we transform the recorded force {axes} to match with the forceps tip {axes}. We calibrate the sensor readings using gravity and tool momentum compensation, previously computed from predefined configurations with the tool attached under no external forces. Finally, we use a scaling value representing {the} force attenuation across the forceps rod. The final value of the force for time $i$ is {given by} Eq.} \ref{eq:force-map}.

{
\begin{equation}
\label{eq:force-map}
F_{EE_{i}} = s_{i} \ast \left(T^{FS_{i}}_{EE_{i}} \times F_{FS_{i}} - \left(T^{FS_{i}^{-1}}_{EE_{i}} \times G_{0} + T_{0} \right) \right)
\end{equation}
}

{where $F_{EE_{i}}$ is the force at the end effector, $s_{i}$ is the force attenuation factor, $T^{FS_{i}}_{EE_{i}}$ is the Cartesian coordinate transformation, $F_{FS_{i}}$ is the force value from the sensor, $G_{0}$ is the moment of gravity compensation and $M_{0}$ is the moment of the tool.}

\section{{Data processing \& Neural Network training}}
\subsection{Mixing datasets towards generalization}
\label{sub:mixing}

\begin{table}[t]
\caption{{Main features} of the used datasets}
\centering
\label{tab:data}
\begin{tabular}{|
>{\columncolor[HTML]{FFFFFF}}l |c|c|}
\hline
\textbf{Dataset} & \cellcolor[HTML]{CCC1DA}\textbf{DaFoEs} & \cellcolor[HTML]{FCD5B5}\textbf{dVRK forceps\cite{chua_toward_2021}} \\ \hline
\textbf{\begin{tabular}[c]{@{}l@{}}Number \\ of samples\end{tabular}} & 71939 & 165380 \\
\textbf{\begin{tabular}[c]{@{}l@{}}Number \\ of scenes\end{tabular}} & 90 & 47 \\
\textbf{\begin{tabular}[c]{@{}l@{}}Predominant \\ task\end{tabular}} & \begin{tabular}[c]{@{}c@{}}Palpation and \\ pushing\end{tabular} & \begin{tabular}[c]{@{}c@{}}Grasping and \\ pulling\end{tabular} \\
\textbf{\begin{tabular}[c]{@{}l@{}}Silicone \\ materials\end{tabular}} & \begin{tabular}[c]{@{}c@{}}2 (Ecoflex-30 \\ and \\ DragonSkin20)\end{tabular} & \begin{tabular}[c]{@{}c@{}}2 (Limbs \& Things \\ phantom \\ and DragonSkin10)\end{tabular} \\
\textbf{\begin{tabular}[c]{@{}l@{}}Tissue \\ structures\end{tabular}} & \begin{tabular}[c]{@{}c@{}}6 (2 single-layer \\ and \\ 4 double-layer)\end{tabular} & \begin{tabular}[c]{@{}c@{}}2 (only double layer \\ considered)\end{tabular} \\
\textbf{Positions} & 5 (planar axis) & \begin{tabular}[c]{@{}c@{}}21 (7 planar and\\ 3 heights)\end{tabular} \\ \cline{1-3}
\end{tabular}
\end{table}

{The capability of predicting forces from tool-tissue interaction using computer vision can vary a lot depending on a large amount of factors: tissue properties (geometry, stiffness or structure), tool velocity/acceleration, type of tool or {and, therefore, the type of} force (pushing, grasping, pulling, etc.). Consequently, each dataset contains each own limitation and biases\cite{torralba_unbiased_2011}. Training in a single dataset leads to good results into the split of the same data used for testing, but may show limited capability in the generalization of the problem. Additionally, datasets for this task are small compared to the ones available for other computer vision problems. Instead, we propose a combination of datasets for training, known as mixing datasets, which has demonstrated effective in other computer vision tasks such as depth estimation \cite{ranftl_towards_2020}.}

{In order to cover many of the variables discussed on the previous paragraph, we design our dataset (DaFoEs) presented in Sec. \ref{sub:dafoes} as a complementary dataset to a previously validated daVinci Research Kit (dVRK) dataset from Chua's paper\cite{chua_toward_2021}. Thus, increasing the volume of data available for training and the variability of such data. And, moreover, exploring the capabilities of such approach on the generalization of vision-based force estimation. We present a short summary of the most meaningful features from each of the two used datasets in Tab. \ref{tab:data}.}

{{In order to cover some missing aspects on dVRK,} we design DaFoEs {as a supporting dataset}. Regarding tissue structure, previous dataset contains only a double layer structure, we {alternate between double and} single layer structures. {These double layer structures could be either a} heterogeneous {or} homogeneous combination of {silicones}. {In addition,} as {the mechanical properties of our} silicone were different {from dVRK}, we increased the total number of materials on the dataset to a total of 4. As dVRK robotic system provided a more reliable grasping and pulling capability, we centred our {experiments} on pushing and palpation, with varying intensity and duration.}

\begin{figure}[t]
    \centering
    \includegraphics[width=0.48\textwidth]{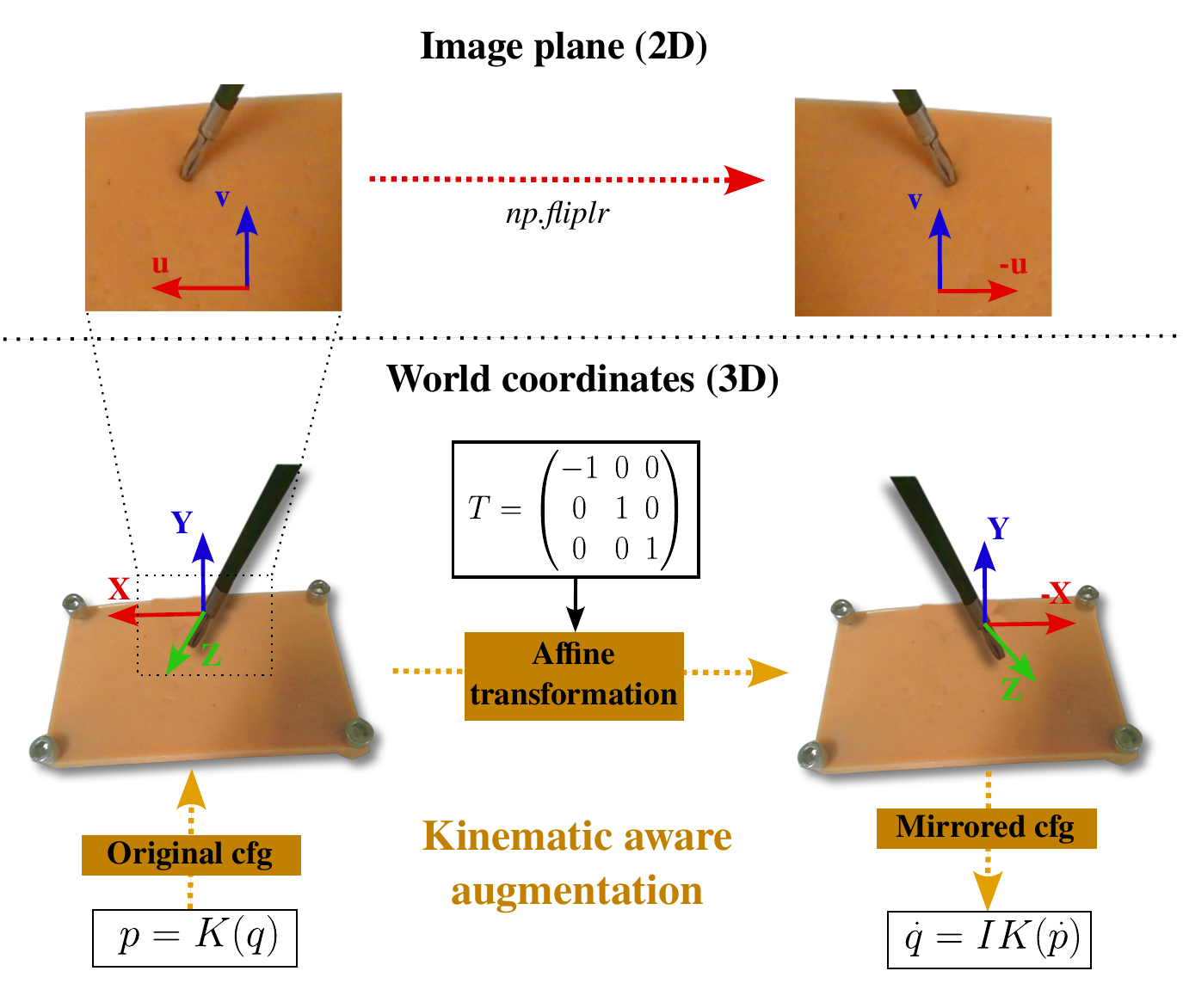}
    \caption{Example for horizontal mirroring transformation for our kinematic aware augmentation pipeline. In the image plane we have the visual transformation. In the lower part we have all the steps to update the kinematic vector of the robot. K stands for kinematics and IK stands for inverse kinematics.}
    \label{fig:augmentation}
\end{figure}

\subsection{Data {generalization} \& augmentation}

{Nevertheless, combining datasets presents a difficult challenge as each of the dataset contains its own individual characteristics. This generates different inputs from different origins and, thus, different features such as image quality, type of camera used for recording, camera viewpoint, state vector size, robotic system used for the interaction. Consequently, in this section we present a data processing pipeline which targets the problem of generalizing the input for dataset mixing, in order to reach a common domain for the solution of our problem.}

\subsubsection{Robot state vector {generalization}}

{We consider the robot state vector to be a valuable piece of information for the prediction of forces. As we do not want to loose any information from the dVRK state vector, we decided to generalized the vector as a 54 element vector, following the representation from Chua {et al.}\cite{chua_toward_2021}. In order to extend our vector to the given size, we {added} the linear and angular velocity, {which is} the first-order derivative approximation of the position and orientation with respect to time. For the rest of the variables, we zero-padded the vectors so they match in length, for example the extra 7th degree of freedom (DOF) from the dVRK.} Lastly, in order to have comparable results, we normalized the state vector subtracting the mean and dividing by the standard deviation of the {combined} training set. 

\subsubsection{Visual data {generalization}}

As we have different {image} acquisition hardware for the visual data {across} the two datasets, we need to define some common pre-processing pipeline for {every image} so {we have a generalized input for the model}. We first centre cropped the images to a square shape of $300\times300$. As the commercial laparoscopic forceps is smaller than the {dVRK} tool, we {further} zoom the image {towards the centroid of the image plane, close to the tool tip, without changing} the {previous} square shape for further processing. We resize the image to $256\times256$. Lastly, we normalized all the images using the ImageNet dataset mean and standard deviation.

\subsubsection{Visual and state data augmentation}

{After inspecting both datasets,} one of the main limitation {that both present} is that we always {approach the tissue for interaction} from the {right} side of the camera plane. {In order to counteract this effect, we designed what we decided to call} Kinematic Aware Augmentations. {These include different linear} transformations on the image plane: rotation ($-20^{\circ} \text{ to } 20^{\circ}$) {and flips around the} vertical and horizontal {axes}. The main difference with {previous augmentation work} that only consider the single {visual data}  stream {is that} in our case {linear transformations {on the image plane also} affect the current position and orientation of the end effector {on the 3D space}.} {Consequently, every time we apply this type of transformations onto the visual data}, we apply the {equivalent three dimensional} affine transformation {{to the related} robot state vector}. In order to transform the orientation, we first transform {the quaternions} into the equivalent rotation matrix{, we apply the transformation in the matrix form and, then, we transform them back into quaternions.} Once the new position of the tool {has been calculated}, we {estimate one of the possible configuration of the robot joints using} Inverse Kinematics (IK)\cite{kazanzides_open-source_2014}. Refer to Figure \ref{fig:augmentation} for a more detailed depiction of the transformation and a {simple} visual example.

{Lastly, as} vision-transformers are sensible to {task-independent} image augmentations \cite{longpre_how_2020}, we decided to include some additional common augmentations {as a way to} improve the performance. These transformations are: random zooming {and} cropping; and hue based transformation (brightness and contrast).

\subsection{{Vision-state} force estimation {using} deep learning}
\label{sub:deep}

\begin{figure}[t]
    \centering
    \includegraphics[width=0.485\textwidth, keepaspectratio]{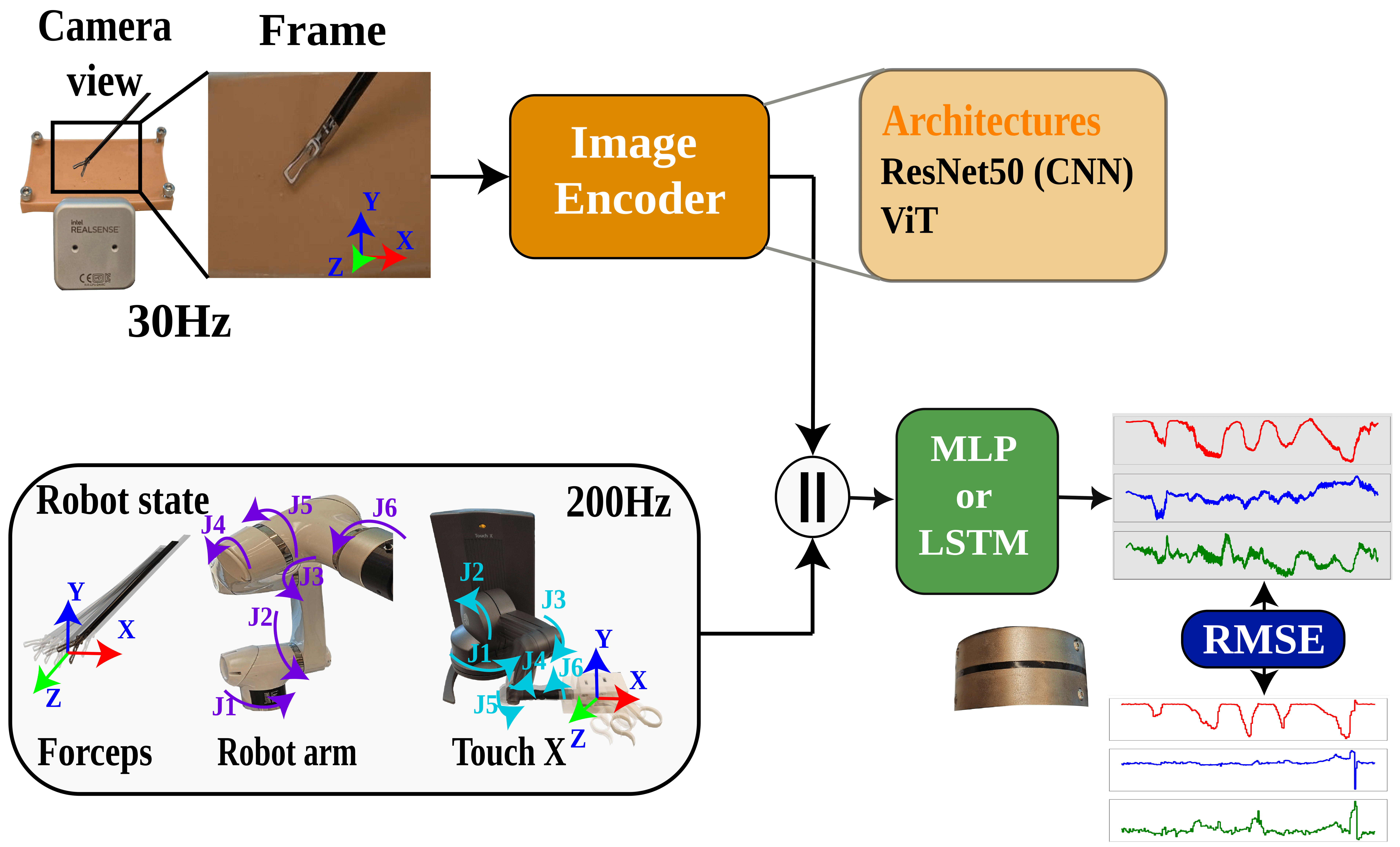}
    \caption{Graphical representation of {our training pipeline for the vision-state models}. In the top right part, we {show the different visual encoders that we used for this research (ResNet50 and Vision Transformer). After concatenation with the state vector, we have the two different types of decoders non-recurrent (MLP) or recurrent (LSTM).}}
    \label{fig:model}
\end{figure}

For this research, {we present vision-state based force ($Y_{i}$) estimation as a regression problem to find a non-linear function $F(.)$ that maps the video frames $X^{\text{video}}_{i}$ and the robot state vector $\rho_{i}$ into a 3-D Cartesian force\cite{marban_recurrent_2019}. In our approach, the function is represented as a deep neural network {with weights $W$}.} {The {architecture} {$F(.)$} is formed by two different parts: {encoder and decoder}. From a general point of view, our method uses the encoder to process solely visual inputs into features in a latent space of higher dimension. In the second step, we concatenate the state vector to the high dimension features, and this combination is processed by the decoder. The method to concatenate the state vector varies depending on the chosen decoder {multilayer perceptron (MLP) or Long-Short Term Memory (LSTM) based recurrent model}. On the other hand, the size of the latent space depends on the encoder (ResNet50 or ViT). A short graphical representation of our architecture can be found at Fig. \ref{fig:model}.}

{{Our MLP decoder is based} on the approach from Chua et al. \cite{chua_toward_2021}. The MLP is formed by dense layers of 84, 180, 50 and 3 channels, with ReLU activation and batch normalization between them. {Additionally, we test this decoder architecture on its own with the name FC (Fully-Connected).}. Based on {literature} \cite{chua_toward_2021, sabique_investigating_2023}, we have the combination between {ResNet50 encoder and MLP. The input for the encoder will be a single image.} Then, we concatenate the robot state vector onto the channel dimension of the {latent}feature space. {Afterwards, these new features are decoded using the MLP. This architecture is further referred to as CNN.}}

{Regarding our second encoder, } the specific architecture used to test the transformer {based image encoder} it is a modified Visual Transformer {from the original paper}\cite{dosovitskiy_image_2021}. The {last layer} tokens are {simplified} using the CLS token reduction method, this means that we use the first token {for predictions}. We substitute the {original} MLP head {for the {common MLP decoder on this paper.} decoder on the convolutional network and the state only network. We concatenate the state vector with the latent features {on the same way as for CNN. This model is further referred to as ViT}.}

{Unlike previous recurrent neural network approaches to force estimation in MIRS \cite{marban_recurrent_2019}, we decide to limit the recurrency size of the network to a sequence of 5 frames. This window corresponds to the minimum amount of state vectors to represent the acceleration of the tool, as the second derivative of its position. Additionally, it considers most of the benefits of LSTM, as they provide better performances on shorter time series, in which the influence from initial inputs is not lost through the different forward inferences. {Our} recurrent {decoder is formed by} 2 blocks of LSTM and a dense layer with 3 channels as output. {We use a different concatenation technique, taking advantage of the extra temporal dimension. We } concatenate them on the temporal dimension after zero-padding the state vector to the same size as the feature vector, making a total of 10 values on this dimension (5 from the frames and 5 from the state vector), similar to {Liu et al.}\cite{liu_camera_2022}}. {To differentiate between the two decoders, we are using the prefix R to name the recurrent models, these models are further referred as RCNN for the ResNet encoder and RViT for the transformer based encoder.}

We train {every} model for 100 epochs. {For weight optimization, we use} Adam optimizer with learning rate of $2\times10^{-4}$ and betas of $(0.9, 0.999)$. {For our loss, we calculate the} Mean Square Error (MSE) between the real value of the sensor and the predicted force from the network to train the model. {In order to accelerate the training and constraint the weights values to the range of our application,} we add L1 normalization during the training. We use a NVIDIA DGX A100 with 8 GPUs {as our hardware for training acceleration.}

\section{Results}
\label{sec:results}

For the testing and comparison of each of the models we use RMSE (Root Mean Square Error), {between the real force from our labels and the predicted force from the model}. As the focus of this paper is {to test} these models for {near real-time} force estimation during {MIRS}, we decided to constrain our metrics to a single task. {This means that we isolated a single clip from each of our datasets, making a total of 2 clips (around 4,500 frames), that were not seen by the model during training, to calculate all the metrics. {The test} video {varies depending on the training type (defined on the next paragraph)}. This method will help us to highlight the strengths and weaknesses of our method for its real life application.}

\begin{figure}[t]
    \centering
    \includegraphics[width=0.45\textwidth, keepaspectratio]{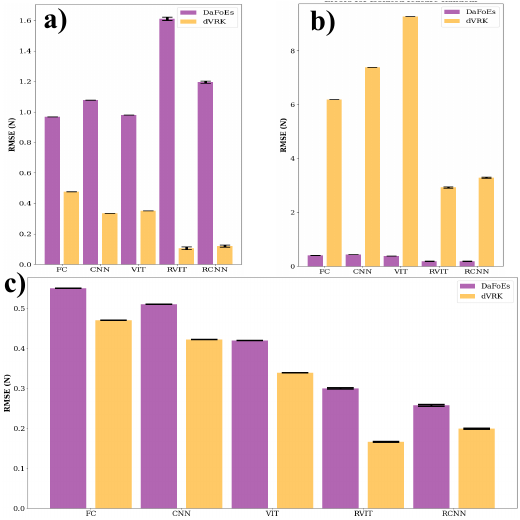}
    \caption{Metrics to compare the effectiveness of our dataset mixing approach. The bars represent the dataset of origin of the testing clip. a) and b) represents the isolated training into a single dataset dVRK and DaFoEs respectively{, and the translation experiment to the opposite dataset}. c) Shows the force difference for the mixed dataset training.}
    \label{fig:mixing}
\end{figure}

{Firstly, we demonstrate the effectiveness of our datasets mixing approach and the data generalization pipeline. We use a \textbf{Random {Rand}} training set up to capture the variability of our mixed dataset. We take the weights for the isolated training (dVRK and DaFoEs) and the mixed approach. For the comparison, we pass our test set 5 times through our trained model and compute the mean and standard deviation of {error}. We present the results using bar plots on Fig. \ref{fig:mixing}. We manage to obtain reliable force estimation with difference lower than $0.2N$ in our best performing recurrent models.} 

\begin{figure}[t]
    \centering
    \includegraphics[width=0.45\textwidth]{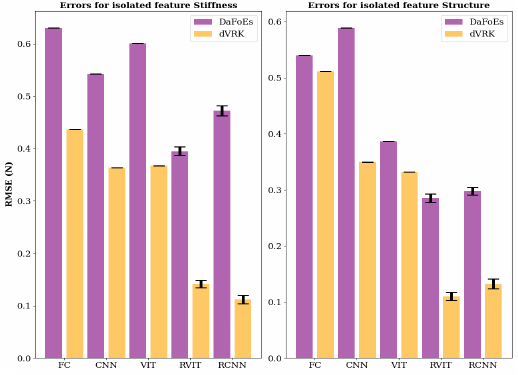}
    \caption{Results for the feature isolation experiment {as bar plots. The X axis shows the different models presented in the paper.}}
    \label{fig:features}
\end{figure}

{On the second part of our analysis, we want to investigate the {weight of certain} feature on both the visual input and the state vector {for the prediction of force, similar to \cite{chua_toward_2021, sabique_investigating_2023}}. For this {part,} we only consider the mixed training scheme as it showed to be the method with less bias and overfitting. For the visual input, we decided to isolate given features that {affect the image information} (refer to Fig. \ref{fig:full_setup} for more information) {such as} \textbf{Stiffness {(Stiff)}} and \textbf{Structure ({Struc)}}. Starting with the isolated \textbf{Stiffness} training, we have 4 different stiffness in total, so we train on 1 stiffness from each dataset and test on the other for each of the dataset. On the other hand, for \textbf{Structure}, the variability is much lower with the majority of the images containing double layer structures. Consequently, we train using double layer geometries for both dataset and, tested onto the single layer structures from the DaFoEs dataset. See the results of this analysis at Fig. \ref{fig:features}.}

\begin{figure}[t]
    \centering
    \includegraphics[width=0.45\textwidth]{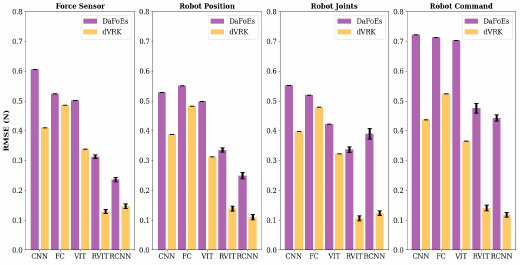}
    \caption{Results for parameter occlusion experiment {as bar plots. The results are presented following the same structure as Fig. \ref{fig:features}.}}
    \label{fig:param}
\end{figure}

{{For} the robot state vector, we identify four variables to have the higher impact onto the prediction of forces: force sensor {(FS)} (specifically for dVRK), robot position {(RP)} , robot joints {(RQ)}  and robot command {(RC)} . When we occlude a state parameter we set the value of such feature to 0, in order to reduce the encoded information our model has for training. For testing, we use the same clips as for the \textbf{Random} training. The results for each of the occluded parameters can be found Fig. \ref{fig:param}.}

\begin{figure*}[t]
    \centering
    \includegraphics[width=0.95\textwidth, height=7cm]{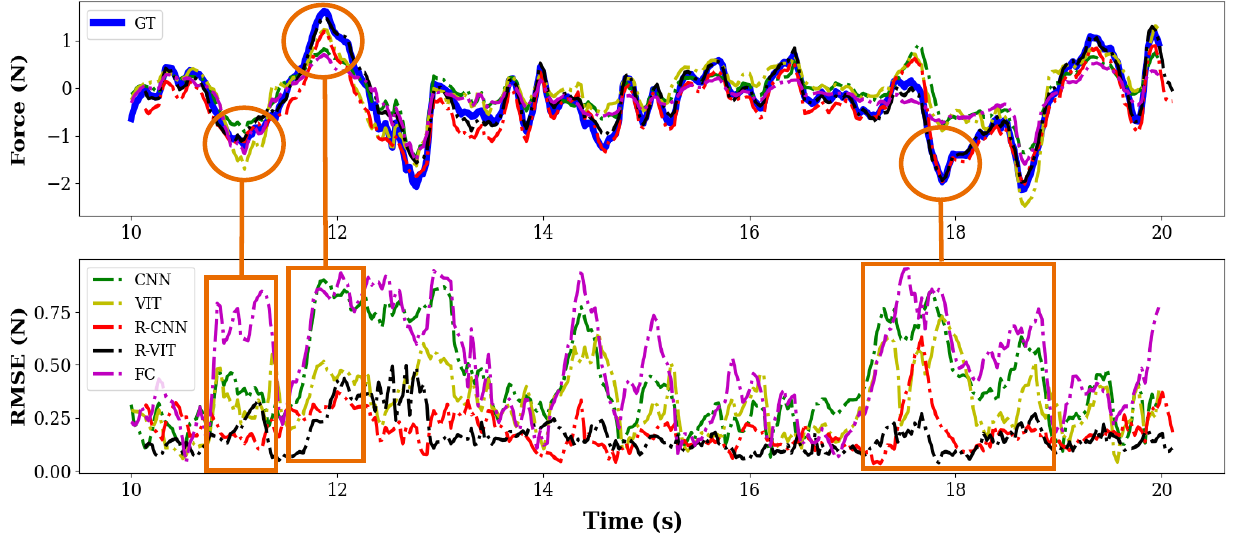}
    \caption{Graph with the evolution of forces {(top)} and the evolution of errors over time {(bottom)}. The graph on the top shows the {temporal evolution of the forces on the X-axis (GT refers to the ground truth value). The graph below shows the temporal change of the RMSE for the 5 different models presented on this paper: Convolutional Neural Network (CNN), Vision Transformer (ViT) and Multilayer perceptron (FC), for both non-recurrent and recurrent cases (R-).}}
    \label{fig:evolution}
\end{figure*}

{As a last experiment and to verify the real effectiveness of our propose recurrent models, we want to observe the temporal evolution of the errors. For this we {compare} the {change of} RMSE {with respect to} time {for the} different architectures, you can find the {plots for the dVRK test clip} in Fig. \ref{fig:evolution}. Going } into the details and focusing onto {the places were {most of the error changes,} we can see that non recurrent models fail} close to force limits. {For this reason,} we isolate peaks (maxima and minima) that are at least 150 readings apart from each other{, 5 seconds,} and we calculate the {mean error} values on those points. The results are collected in Table \ref{tab:peaks}. 

Lastly, we compute the inference time for each of the model considering the forward pass through the model {, excluding pre-processing}. We simulate 1000 passes and we record the mean time in seconds {to} compute the frequency (Hz) using this value. Both CNN and ViT show similar {inference frequency} ($80$Hz), {enough} for near real-time applications. Whereas, recurrent models have {a} higher inference time of around $12.45$Hz.

\setlength\extrarowheight{2.5pt}
\begin{table*}[t]
\caption{{Error values at the isolated local maxima and minima of the force. They are at least 5 seconds apart from each other}}
\centering
\label{tab:peaks}
\hspace*{-0.8cm}\begin{tabular}{|llccccccccccccccc|}
\hline
\multicolumn{17}{|c|}{\textbf{RMSE (N)}} \\ \hline
\multicolumn{2}{|c|}{} & \multicolumn{5}{c|}{\cellcolor[HTML]{FCD5B5}\textbf{dVRK\cite{chua_toward_2021}}} & \multicolumn{5}{c|}{\cellcolor[HTML]{CCC1DA}\textbf{DaFoEs}} & \multicolumn{5}{c|}{\cellcolor[HTML]{B9CDE5}\textbf{Mixed}} \\ \cline{3-17} 
\multicolumn{2}{|c|}{\multirow{-2}{*}{\textbf{Training mode}}} & \textbf{MLP} & \textbf{CNN} & \textbf{RCNN} & \textbf{ViT} & \multicolumn{1}{c|}{\textbf{RViT}} & \textbf{MLP} & \textbf{CNN} & \textbf{RCNN} & \textbf{ViT} & \multicolumn{1}{c|}{\textbf{RViT}} & \textbf{MLP} & \textbf{CNN} & \textbf{RCNN} & \textbf{ViT} & \textbf{RViT} \\ \hline
\multicolumn{1}{|l|}{} & \multicolumn{1}{l|}{\textbf{Rand}} & 0.91 & 0.56 & 0.16 & 0.65 & \multicolumn{1}{c|}{\textbf{0.14}} & 1.48 & 1.64 & 0.44 & 0.95 & \multicolumn{1}{c|}{\textbf{0.34}} & 0.84 & 0.64 & \textbf{0.18} & 0.47 & 0.28 \\
\multicolumn{1}{|l|}{} & \multicolumn{1}{l|}{\textbf{Stiff}} & 0.91 & 0.56 & 0.16 & 0.56 & \multicolumn{1}{c|}{\textbf{0.12}} & 1.35 & 1.47 & \textbf{0.55} & 0.88 & \multicolumn{1}{c|}{0.58} & 0.72 & 0.56 & \textbf{0.16} & 0.55 & 0.18 \\
\multicolumn{1}{|l|}{\multirow{-3}{*}{\textbf{Features}}} & \multicolumn{1}{l|}{\textbf{Struc}} & 0.65 & 0.56 & 0.15 & 0.56 & \multicolumn{1}{c|}{\textbf{0.13}} & 1.76 & 0.64 & 0.42 & 0.99 & \multicolumn{1}{c|}{\textbf{0.30}} & 0.91 & 0.55 & 0.17 & 0.54 & \textbf{0.14} \\ \hline
\multicolumn{1}{|l|}{} & \multicolumn{1}{l|}{\textbf{FS}} & 0.67 & 0.55 & 0.14 & 0.57 & \multicolumn{1}{c|}{\textbf{0.12}} & 1.58 & 1.05 & 0.35 & 0.95 & \multicolumn{1}{c|}{\textbf{0.30}} & 0.80 & 0.76 & 0.19 & 0.50 & \textbf{0.16} \\
\multicolumn{1}{|l|}{} & \multicolumn{1}{l|}{\textbf{RP}} & 0.72 & 0.66 & 0.16 & 0.65 & \multicolumn{1}{c|}{\textbf{0.12}} & 0.61 & 1.24 & 0.38 & 0.80 & \multicolumn{1}{c|}{\textbf{0.30}} & 0.83 & 0.63 & \textbf{0.13} & 0.47 & 0.19 \\
\multicolumn{1}{|l|}{} & \multicolumn{1}{l|}{\textbf{RQ}} & 0.75 & 0.55 & 0.15 & 0.47 & \multicolumn{1}{c|}{\textbf{0.14}} & 1.31 & 1.49 & 0.36 & 0.72 & \multicolumn{1}{c|}{\textbf{0.24}} & 0.81 & 0.57 & 0.16 & 0.46 & \textbf{0.13} \\
\multicolumn{1}{|l|}{\multirow{-4}{*}{\textbf{Occlusion}}} & \multicolumn{1}{l|}{\textbf{RC}} & 0.64 & 0.64 & 0.15 & 0.47 & \multicolumn{1}{c|}{\textbf{0.13}} & 2.82 & 2.38 & \textbf{0.81} & 2.31 & \multicolumn{1}{c|}{1.26} & 0.86 & 0.65 & \textbf{0.16} & 0.51 & 0.21 \\ \hline
\end{tabular}\hspace*{-0.8cm}
\end{table*}

\section{Discussion}

The results from Fig. \ref{fig:mixing} show that isolated training overfits to the current {data domain}. {It can be seen that results on their testing split are good, but they lack from information to generalize the prediction to the second set}. Most clearly when we try to translate from DaFoEs to dVRK the model shows an increase in error of $2,218\%$, due to the {more complete state vector from the robot} and {different camera recording angle} used in this dataset. Nevertheless, using this simpler dataset {as a secondary dataset to reduce bias to robotic system, workspace and image domain} {towards} generalization does provide good results. However, the data available is still not large enough to drive a closing solution for the generalization task, {at least an additional dataset with different hardware for the interaction and video recording to test our generalization approach {would be} preferable}.

{Regarding the deep learning training,} across all the {trained models,} we can {conclude} that the recurrent models, due to their temporal expansion, do perform {significantly} better than the rest of the models. {Nonetheless, these models do present higher variability during the 5 runs we have done through our test set in the three of the training modes {coming mainly from the decoder}.} The improvement of such architectures varies depending on the encoder and type of training, {highlighting the Vision Transformer based encoder with a RMSE as low as $0.17N$ of force variation}. Even though, ViT based models perform worse across domains, we see a minimal improvement in performance when {using our dataset mixing pipeline for generalization}. The performance of the transformer increases up to $16\%$ for dVRK dataset and $59\%$ for the newly presented DaFoEs.

{Starting from the transfer of visual information on our second experiment, we can observe that most of the models had a slight increase on performance. Isolating a single feature does actually reduce the volume of training data, therefore making the model be more specified. For example for RViT we have a values of $0.17N$, $0.14N$ and $0.11N$, respectively for Random, Stiffness and Structure. We know from our discussion in Sec. \ref{sec:results} that structure contains a larger volume of images as most of the images are double layer. Even though we can see that specific training could be a good approach for these models, more experiments are needed to {drive} any conclusion.}

{For parameter occlusion we have a similar behavior, in which blocking the network to a certain information during training makes the network perform slightly better on the test set. From all the features, we can conclude that force sensor readings from dVRK are the one encoding the most information for non-recurrent models with an improvement of $5\%$. However, specifically for recurrent models robot command has the most information as the models are only improved for around $12\%$. Additionally, recurrent models present a much higher variability across inferences when parameters are occluded passes from $0.001N$ to a maximum of $0.01N$ for the robot command using the RViT architecture.}

As we can observe from the results in Tab. \ref{tab:peaks} {and in Fig. \ref{fig:evolution}}, {non-recurrent} force prediction models do suffer close to local maxima and minima {where there is an increase of force, as they lack from the context coming from previous readings}. The increase of error is clear for the fully connected kinematic alone, with an increase of $47\%$ compared to the mean error. As from previous discussion, we can see that the recurrent models do keep an almost constant error even at these peaks with a minimal increase of $0.05$N and $0.04$N for RViT and RCNN respectively, {meaning that most of their error is coming from steady states when there is no contact}. Nonetheless, inference time is a big concern for the {use} of such networks on near-real-time {implementation for} MIRS. In this case, we have a residual difference between the {different encoders. However, the decoders present a large difference in inference time}. This means that {there is a necessity to find a way to optimize the inference these models \cite{liu_camera_2022}}. 

{The collection of data for {sensorless force estimation} presents many challenges on its own. Making a reliable set up {, deciding the phantoms or animal tissue to model human in-vivo soft environments, for example,} and {the} control unit for a laparoscopic tool is not a trivial challenge. Moreover, it is hard to decide which data from the state vector {should be prioritized}, as {they could be} related to each other {via mathematical formulation, for example the acceleration as the second-order derivative of the position or inverse kinematics for joints}. We should also consider that {the teleoperated framework used for the collection} may induce bias on this data. Additionally, the camera position and recording angle have a big influence on the capacity of the model to make reliable predictions. So, when recording new datasets for this task these features should be taken into account.} {For the different models, as force is linearly related to the acceleration of the tool, we conclude that using {an additional temporal channel to decode features} is the optimal approach {for sensorless vision-state based force estimation}. In other words, having the knowledge were the tool has been on previous time steps it is useful for the prediction, {specifically on the phases were higher force is actively applied}. However, it requires from a correct temporal sampling as it has previously been demonstrated \cite{chua_toward_2021, liu_camera_2022, lee_neural_2023, sabique_investigating_2023}. In order to improve our approach, we suggest to explore the effect of recurrency on the encoder, processing the sequence as a single input for example, and explore the possibilities of using different encoding methods of the temporal sequence to feed the decoder.}

\section{Conclusion \& Future work}

In this research, we demonstrate the viability of {datasets mixing for training} different deep neural networks for the {sensorless vision-state force estimation as a possible general approach} in {MIRS}. We show that {using the correct temporal sampling largely improves the performance of the {temporal decoder}. In general, all the models analysed in this paper can learn the tendency of forces, but only recurrent models predict the {full} range of forces all across {a complete clip}. Additionally, for our} mixing dataset {pipeline}, we observe that transformer architectures do benefit from {from the creation} { of larger volumes of data, even though {hardware systems vary for both the state and visual recordings}}. However, we are still at the {initial steps} of this area of research and more dataset, architectures and learning techniques should be developed in order to reach a consensus and expand our knowledge on this topic.

{For this reason,} new research should focus on the collection of {new more variable} dataset{s, to explore this generalization pipeline from multiple and more varied origins. Some examples for the origin of data are:} from simulated environments for large data volumes, complex phantom geometries to have a better modelling of the lumen{, ex-vivo animal or human tissue \cite{lee_learning_2023},} and in-vivo environments for a more realistic visual input. However, these environments do present more problems regarding the use of reliable force sensing hardware due to their confined workspace {making it difficult to make use of supervised training schemes}. Consequently, {there is a necessity to create additional theoretical formulations over which to build new training pipelines that avoid the necessity of ground truth reliable force readings}.

\section{Acknowledgement}
Special thanks to Prof. Zhonge Chua from Case Western Reserve University for sharing their dataset with us for this work.

\bibliographystyle{ieeetr}
\bibliography{bibliography}

\end{document}